\def\year{2020}\relax
\documentclass[letterpaper]{article} 
\usepackage{aaai20}  
\usepackage{times}  
\usepackage{helvet} 
\usepackage{courier}  
\usepackage[hyphens]{url}  
\usepackage{graphicx} 
\graphicspath{ {images/} }
\usepackage{graphicx}  
\usepackage{amsmath}
\usepackage{amssymb}
\title{Quality of syntactic implication of RL-based sentence summarization}
\author{Hoa T. LE \textsuperscript{1}, Christophe Cerisara \textsuperscript{2}, Claire Gardent \textsuperscript{2} \\
  {\tt \textsuperscript{1} LORIA, UMR 7503, Nancy, France; \textsuperscript{2} CNRS-LORIA, France}}
 
\begin{document}

\maketitle

\begin{abstract}
Work on summarization has explored both reinforcement learning (RL)
optimization using ROUGE as a reward and syntax-aware models,
such as models whose input is enriched with 
part-of-speech (POS)-tags and dependency information. However, it is not
clear what is the respective impact of these approaches beyond the standard ROUGE evaluation metric. Especially, RL-based for summarization is becoming more and more popular. In
this paper, we provide a detailed comparison of these two
approaches and of their combination along several dimensions that
relate to the perceived quality of the generated summaries: number of repeated words, distribution of part-of-speech tags, impact of sentence length, relevance and grammaticality.
Using the standard Gigaword sentence summarization task,
we compare an RL self-critical sequence training (SCST)
method with syntax-aware models that leverage POS tags and Dependency information.
We show that on all qualitative evaluations, the combined model gives
the best results, but also that only training with RL and without
any syntactic information already gives nearly as good results as
syntax-aware models with less parameters and faster training convergence.

\end{abstract}

\section{Introduction}

Early neural approaches to text generation tasks such as machine
translation, summarization and image captioning mostly relied on
sequence-to-sequence models \cite{Sutskever2014} where the model was
trained using cross-entropy and features were learned automatically.
More recent work however, shows that using reinforcement
learning or explicitly enriching the input with additional linguistic
features helps to improve performance.

Reinforcement learning was proposed to address two shortcomings of
cross entropy training. First, there is a discrepancy between how the
model is trained (conditioned on the ground truth) and used at test
time (using argmax or beam search), namely the \textit{exposure bias}
problem. Second, the evaluation metrics (for ex. ROUGE,
METEOR, BLEU, etc.) differ from the objective that is maximized with
the standard cross-entropy on each token; this is known as the \textit{loss mismatch}
problem. Typically, RL is used to optimize task-specific objectives
such as ROUGE for text summarization systems
\cite{paulus-socher-arxiv17,N18-1158,N18-1150,pasunuru-bansal:2018:N18-2}
and SARI \cite{wei16tacl}  for 
sentence simplification models \cite{zhang-emnlp17}.


Similarly, while neural networks allow for features to be learned
automatically, explicitly enriching the input with linguistic
features was repeatedly shown to improve performance.
For instance,
\cite{sennrich-haddow-2016-linguistic,li-etal-2017-modeling} show that
adding morphological features, part-of-speech (POS) tags, syntactic
dependency and or parse trees as input features improves the
performance of neural machine translation (NMT) systems; and
\cite{nallapati-etal-2016-abstractive} that integrating linguistic
features such as POS tags and named-entities helps to improve
summarization. 

In this paper, we explore the relative impact of these two approaches
on sentence summarization.
More precisely, we assess and compare the quality of the summaries generated by
syntax-aware, RL-trained and combined models with regard to
several qualitative aspects that strongly impact the perceived
quality of the generated texts: number of repetitions, sentence length,
distribution of part-of-speech tags, relevance and grammaticality.

Using the standard Gigaword benchmark corpus, we
compare and combine an RL self-critical sequence training (SCST) method with
syntax-aware models that leverage POS tags and/or dependency information.
We show that both enhancements, syntactic information and RL training,
benefit to a sequence-to-sequence summarization model with attention and
copy-pointer mechanism.
While the combined model gives the best quality of summaries, we also show that the reinforcement learning
approach alone may be preferred when computational complexity is an issue, as
it gives nearly as good results as the syntax-aware model but with less parameters and
faster training convergence.



\section{Related Work}

We briefly discuss previous work on syntax-aware and RL-based models
for text-to-text generation focusing on summarization and NMT and
we position our work with respect to these approaches.

\vspace{0.2cm}
\noindent\textbf{Syntax models:} Explicit modeling of syntax has frequently been used in text generation applications in particular, for NMT and summarization. Thus \cite{sennrich-haddow-2016-linguistic} enrich the input to NMT with dependency labels, POS tags, subword tags and lemmas so that each input token is represented by the concatenation of word and features embeddings. Similarly, \cite{nallapati-etal-2016-abstractive} enrich the encoder side of a neural summarization model with POS tag, NER tag, TF-IDF features. The intuition is that words will be better disambiguated by taking syntactic context into account. Speculating that full parse trees can be more beneficial for NMT than shallow syntactic information, \cite{li-etal-2017-modeling} enrich the input with a linearization of its parse tree and compares three ways of integrating parse tree and word information (parallel, hierarchical and mixed). Other works have focused on integrating syntax in the decoder or through multi-task learning. Thus, \cite{Le2017ImprovingST} defines machine translation as a sequence-to-dependency task in which the decoder generates both words and a linearized dependency tree while \cite{kiperwasser-ballesteros-2018-scheduled} propose a scheduled multi-task learning approach where the main task is translation but the model is alternatively trained on POS tag, Dependency Tree and translation sequences.

Our model for syntax-aware summarization is similar to \cite{li-etal-2017-modeling} in that we use a hierarchical model to integrate syntax in the encoder. We depart from it in that (i) we enrich the input with POS tag and/or dependency information rather than constituency parse trees; (ii) we apply our model to summarization rather than translation.

\vspace{0.2cm}
\noindent\textbf{RL sequence models:} 
Various RL models have been proposed for sequence-to-sequence models. \cite{DBLP:journals/corr/RanzatoCAZ15} introduce an adaptation of REINFORCE \cite{Williams92simplestatistical} to sequence model and a curriculum learning strategy to alternate between ground truth and the sample from the RL model. This vanilla REINFORCE is known to have high variance. Thus, a learned baseline is equipped to mitigate this issue. \cite{DBLP:journals/corr/BahdanauBXGLPCB16} propose another reinforcement learning model, namely actor-critic, to have lower variance of model estimations, offsetting by a little bias. The impact of the bias, as well as the goodness of the model, relies particularly on the careful design of the \textit{critic}. In practice, to ensure convergence, intricate techniques must be used including an additional target network \textit{Q'}, delayed actor, a critic value penalizing term and reward shaping. In contrast, \cite{Rennie2017SelfCriticalST} introduce a very simple and effective way to construct a better baseline for REINFORCE, namely self-critical sequence training (SCST) method. Instead of looking for and constructing a baseline or using a real critic as above, SCST uses its own prediction normally used at inference time to construct the sequence and uses this to normalize the reward. \cite{KryscinskiPXS18} adapt this training method to improve the abstraction of text summarization via a combination of cross-entropy, policy learning, pretrained language model and novel phrase reward. Similarly, we use SCST to train a summarization model. However, our model uses ROUGE as a reward and we focus on comparing models trained using different learning strategies (RL vs Cross Entropy) and informed by different sources (with and without syntax). 

\section{Models}

We train and compare models that differ along two main dimensions: training (cross-entropy vs. RL) and syntax (with and without syntactic information). 

\subsection{Baseline} The baseline is a sequence-to-sequence model consisting of a bidirectional LSTM encoder and a decoder equipped with and an attention and a copy pointer-generator mechanism.    

\vspace{0.2cm}
\noindent\textbf{Encoder.} The source sentence is encoded using two recurrent neural networks (denoted \textit{bi-RNN}) \cite{Chung2014}:
one reads the whole sequence of words $x = (x_1, ..., x_m)$ from left to right and the other from right to left.
This results in a forward and backward sequence of hidden states $(\overrightarrow{h_1}, ..., \overrightarrow{h_m})$ and $(\overleftarrow{h_1}, ..., \overleftarrow{h_m})$ respectively.
The representation of each source word $x_j$ is the concatenation of the hidden states $h_j = [\overrightarrow{h_j},\overleftarrow{h_j}]$.

\vspace{0.2cm}
\noindent\textbf{Decoder.} An RNN is used to predict the target summary $y=(y_1, ..., y_n)$.
At each decoder timestep, a multi-layer perceptron (MLP) takes as input the recurrent hidden state $s_i$, the previously predicted word $y_{i-1}$ and a source-side context vector $c_i$
to predict the target word $y_i$.
$c_i$ is the weighted sum of the source-side vectors $(h_1, ..., h_m)$.
The weights in this sum are obtained with 
an attention model~\cite{bahdanau+al-2014-nmt}, which computes the similarity between the target vector $s_{i-1}$ and every source vector $h_j$.

\vspace{0.2cm}
\noindent\textbf{Copy-Pointer.} The attention encoder-decoder tends to ignore the presence of rare words in the source,
which might be important especially for the summarization task. The Copy-Pointer~\cite{see-etal-2017-get} enables to either copy source words via a pointer or generate words from a fixed vocabulary.
A soft switch $p_{gen}$ is learned to choose between \textit{generating} a word from the vocabulary by sampling from the output distribution of the decoder, or \textit{copying} a word from the input sequence by sampling from the attention distribution.

\subsection{Integrating Syntax}

\begin{figure}[t!]
\centering
\includegraphics[scale=0.40]{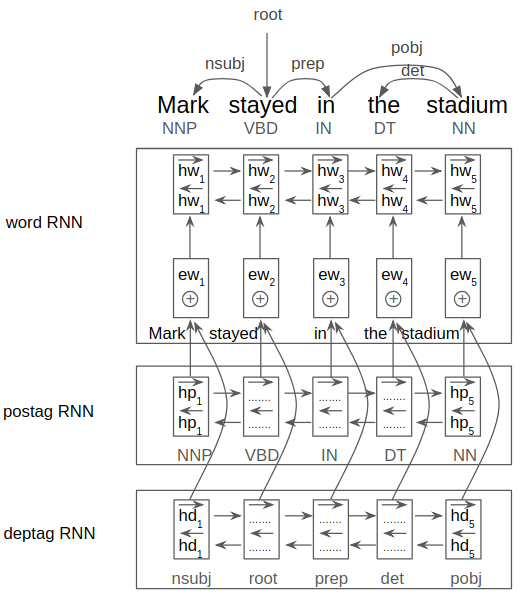}
\footnotesize
\caption{Pos+Deptag Model}
\label{fig:syntax_model}
\end{figure}

Conventional neural summarization models only rely on the sequence of raw words and ignore syntax information.
We include syntactic information in our summarization model using the hierarchical-RNN topology introduced by \cite{li-etal-2017-modeling} and comparing three sources of information: POS tags (Postag), dependency relations (Deptag) and their combination (Pos+Deptag). Figure~\ref{fig:syntax_model} shows a graphical depiction of the Pos+tag model. In essence, each source of information (sequence of tokens, of POS tags, of dependency relations) is encoded using a bidirectional LSTM and each input token is then represented by the concatenation of the hidden-states produced by each  information source considered. For instance, in the Postag model, the POS tag bi-LSTM takes as input a sequence of POS tags and outputs a sequence of hidden states $\left(hp_j=[\overleftarrow{hp_j};\overrightarrow{hp_j}]\right)_{1\leq j\leq m}$ similarly to the word bi-RNN.
Each $hp_j$ is then concatenated with the input word embeddings $ew_j$ and passed on to the word bi-RNN.



For the Deptag model, the input sequence to the Deptag bi-LSTM includes, for each input tokens, the name of the dependency relation that relates this token to its syntactic head (e.g., \textit{nsubj} for the token ``Mark'' in the sentence shown at the top of Figure~\ref{fig:syntax_model}). The Deptag bi-LSTM then output a sequence of hidden states $\left(hd_j=[\overleftarrow{hd_j};\overrightarrow{hd_j}]\right)_{1\leq j\leq m}$ which  
are concatenated with the corresponding word embeddings $ew_j$ and passed to the word bi-RNN.

Finally, for the Pos+Deptag model, both the POS tag and the syntactic hidden states are
concatenated with the words embeddings to give the final input vector $[\overleftarrow{ew_j};\overrightarrow{ew_j};\overleftarrow{hp_j};\overrightarrow{hp_j};\overleftarrow{hd_j};\overrightarrow{hd_j}]$ which  is passed on to the upper-level word bi-RNN.


\subsection{RL Learning}

\paragraph{Summarization as an RL problem}

Neural summarization models are traditionally trained using the cross entropy loss. \cite{DBLP:journals/corr/RanzatoCAZ15} propose to directly optimize Natural Language Processing (NLP)
metrics by casting sequence generation as a Reinforcement Learning problem. As most NLP metrics (BLEU, ROUGE, METEOR,...) are non-differentiable, RL is appropriate to reach this objective.
The parameters $\theta$ of the neural network define a \textit{natural policy} $p_\theta$, which can be used to predict the next word.
At each iteration, the decoder RNN updates its internal state (hidden states, attention weights, copy-pointer weights...).
After generating the whole sequence, a reward $r(\cdot)$ is computed, for instance the ROUGE score.
This reward is evaluated by comparing the generated sequence and the gold sequence. The RL training objective is to minimize the negative expected reward:

\begin{equation}
L^{RL}(\theta)= -  \mathbb{E}_{w^s\sim p_{\theta}} \left[ r(w^s)\right]
\end{equation}     
where $w^s= (w^s_1,\dots w^s_T)$ and  $w^s_{t}$ is the word sampled from the decoder at time step $t$.
Following~\cite{Williams92simplestatistical}, the gradient $\nabla_{\theta}L^{LR}(\theta)$ can be computed as follows:
\begin{equation}
\nabla_{\theta}L^{RL}(\theta)= - \mathbb{E}_{w^s \sim p_{\theta}} \left[ r(w^s) \nabla_{\theta} \log p_{\theta}(w^s) \right]
\end{equation}

In practice, the vanilla REINFORCE yields a very high variance during training.
In order to help the model stabilize and converge to good local optima, vanilla REINFORCE is extended to compute the reward \textit{relative} to a \textit{baseline} $b$:
\begin{equation}
\begin{aligned}
& \nabla_{\theta}L^{RL}(\theta)= \\
& - \mathbb{E}_{w^s \sim p_{\theta}} \left[ (r(w^s)- b ) \nabla_{\theta} \log p_{\theta}(w^s)\right]
\end{aligned}
\label{eq:baseline_rl}
\end{equation}
This baseline can be an arbitrary function (function of $\theta$ or $t$), as long as it does not depend on $w^s$ \cite{sutton1998rli}.

\paragraph{Self-critical sequence training}

There are various ways to reduce RL variance and choose a proper baseline: for instance, using a second decoder~\cite{DBLP:journals/corr/RanzatoCAZ15}
or building a \textit{critic network} and optimizing with a \textit{value network} instead of real reward~\cite{DBLP:journals/corr/BahdanauBXGLPCB16}.
In the following, we have chosen the self-critical sequence training (SCST) technique~\cite{Rennie2017SelfCriticalST}, which has been shown to be very simple and effective.
The main idea of SCST is to use, as baseline in the vanilla REINFORCE algorithm, the reward obtained with the inference algorithm used at test time.
Equation~\ref{eq:baseline_rl} then becomes:
\begin{equation}
\begin{aligned}
& \nabla_{\theta}L^{RL}(\theta) = \\
& - \mathbb{E}_{w^s \sim p_{\theta}} \left[ (r(w^s)- r(\hat{w}) ) \nabla_{\theta} \log p_{\theta}(w^s)\right] \\
\end{aligned}
\end{equation}
where $r(\hat{w})$ is the reward obtained by the current model with the inference algorithm used at test time.
As demonstrated by \cite{Zaremba2015ReinforcementLN}, we can rewrite this gradient formula as:
\begin{equation}
\begin{aligned}
& \frac{\partial L^{RL}(\theta)}{\partial s_{t}} = \\
& (r(w^s)-r(\hat{w}))(p_{\theta}(w_{t}|w_{t-1}^s,h_t)- 1_{w^s_{t}})
\end{aligned}
\end{equation}
where $s_t$ is the input to the final softmax function in the decoder.
The term on the right side resembles logistic regression, except that the ground truth $w_t$ is replaced by sampling $w^s_{t}$.
In logistic regression, the gradient is the difference between the prediction and the actual 1-of-N representation of the target word:
\begin{equation}
\begin{aligned}
\frac{\partial L^{XENT}(\theta)}{\partial s_{t}} = p_{\theta}(w_{t}|w_{t-1},h_t)- 1_{w_{t}}
\end{aligned}
\end{equation}
We see that samples that return a higher reward than $r(\hat{w})$ will be encouraged while samples that result in a lower reward will be discouraged.
Therefore, SCST intuitively tackles well the exposure bias problem as it forces to improve the performance of the model with the inference algorithm used at test time.
In order to speed up sequence evaluation at training time, we use greedy decoding with $\hat{w}_t = \arg \max_{w_t} p(w_t\,|\,h_t)$.
\footnote{Two variants of this training method exist: TD-SCST and the ``True'' SCST, but both variants do not lead to significant additional gain on image captioning~\cite{Rennie2017SelfCriticalST}.
So we didn't explore these two variants for summarization as greedy decoding already obtains quite good result. We leave this for future work.}

\paragraph{Training objective and Reward}

The number of words in the vocabulary may be quite large in text generation, which leads to a large state space that 
may be challenging for reinforcement learning to explore.
To reduce this effect, we follow~\cite{KryscinskiPXS18} and adopt a final loss that is a linear combination of
the cross-entropy loss and the policy learning loss:
\begin{equation}
\begin{aligned}
L = (1-\alpha) L^{XENT} + \alpha L^{RL}
\end{aligned}
\label{eq:loss}
\end{equation}
$\alpha$ is a hyper-parameter that is tuned on the development set.

We use ROUGE-$F_1$  as the reward for the reinforce agent as the generation should be as concise as the gold target sentence.

\section{Experiments}

\paragraph{Data}

We evaluate our approach on the Gigaword corpus \cite{Rush15}, a
corpus of 3.8M sentence-headline pairs and where the average input
sentence length is 31.4 words (in the training corpus) and the average
summary length is 8.3 words. The test set consists of 1951 sentence/summary pairs. As~\cite{Rush15}, we use 2000 sample pairs (among 189K pairs) as development set.

\paragraph{Automatic Evaluation Metric}
	We adopt ROUGE~\cite{lin2004rouge} for automatic evaluation.	
	It measures the quality of the summary by computing overlapping lexical units between the candidate and gold summaries.
	We report ROUGE-1 (unigram), ROUGE-2 (bi-gram) and ROUGE-L (longest common sequence) F1 scores.
	ROUGE-1 and ROUGE-2 mainly represent informativeness while ROUGE-L is supposed to be related to readability \cite{DBLP:conf/aaai/CaoWLL18}.

\paragraph{Implementation}
Our models implementations are based on the Fast-Abs-RL~\cite{chen2018fast} code~\footnote{\scriptsize{\url{https://github.com/ChenRocks/fast_abs_rl}}}.
Although this code is not optimized to give the best possible performances,
it is flexible enough to allow for the integration of syntactic features.

\begin{figure}[th!]
\centering
\includegraphics[scale=0.48]{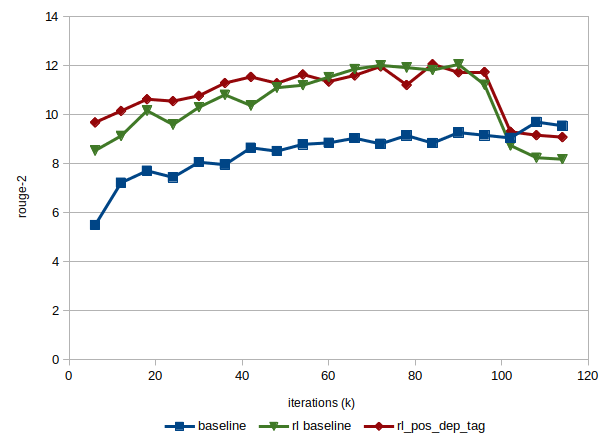}
\footnotesize
\caption{Catastrophic forgetting of the RL decoder on the Gigaword dev set}
\label{fig:ablation_rl}
\end{figure}

The hyperparameter $\alpha$ in Eq~\ref{eq:loss} needs careful tuning. Figure~\ref{fig:ablation_rl} illustrates 
a problematic case when $\alpha$ continuously increases until it reaches $\alpha=1$ at iteration $10^5$, where the RL models forget the previously learned patterns and degenerate.
A good balance between exploration and supervised learning is thus necessary to prevent such catastrophic forgetting.
We have found on the development set that 
the Reinforcement Learning weight $\alpha$ may increase linearly by (step/$10^5$) with the number of training iterations, until it reaches
a maximum of 0.82 for the RL-s2s model and of 0.4 for the RL-s2s-syntax model.

For all models, we use the Adam optimizer~\cite{kingma2014adam} with a learning rate of 0.001 (tuned on the dev set).
The word vocabulary size is 30k, number of part-of-speech tags 40 and number of dependency tags 244.
We have chosen the default size (from the codebase) of 128 for word embeddings, and arbitrarily 30 dimensions both for the part-of-speech and dependency embeddings.
Similarly, we have chosen the default values of 256 hidden states for every bidirectional RNN,
32 samples for the batch size, a gradient clipping of 2 and early-stopping on the development set.
Our adapted code is given as supplementary material and will be published with an open-source licence.

\begin{table*}[h]
\centering
\begin{tabular}{l|c|c|c|c|c}
\hline \bf Models & \bf \#Params & \bf Time for 1 epoch & \bf R-1 & \bf R-2 & \bf R-L \\ \hline
Re3Sum \cite{cao-etal-2018-retrieve} & - & - & 37.04 & 19.03 & 34.46 \\
\hline
Our Baseline s2s & 6.617M & 13h18m & 27.57 & 10.29 & 26.02 \\
\hline
Postag s2s & +312k & +54m & 30.52 & 12.13 & 28.8 \\
Deptag s2s & +318k & +1h13m & 30.25 & 12.15 & 28.73 \\
Pos+Deptag s2s & +630k & +1h34m & 30.8 & 12.3 & 29.22 \\
\hline
RL s2s & +0 & -1h35m & 29.94 & 11.64 & 28.54 \\
RL postag s2s & - & - & 30.82 & 12.19 & 29.12 \\
RL deptag s2s & - & - & 30.58 & 12.08 & 29.01  \\
RL pos+deptag s2s & - & - & 30.76 & 12.31 & 29.11 \\
\hline
\end{tabular}
\caption{Performance comparisons between models}\label{tab:main_results}
\end{table*}

\section{Results and Analysis}

\paragraph{ROUGE.} Table~\ref{tab:main_results} shows the performances measured in ROUGE score. State-of-the-art summarization system come from \cite{cao-etal-2018-retrieve}, which appears to be the best system on the Gigaword corpus reported in \url{http://nlpprogress.com/english/summarization.html}, as of May 2019.

Both Syntactic and RL models outperform the baseline.

Syntax-aware models outperform the baseline by +1.84 (Postag), +1.86  (Deptag) and  +2.01 (Dep+Postag) ROUGE-2 points.

While RL without syntax slightly under-performs syntactic models, it still achieves an improvement of +1.35 rouge-2 over the baseline. In other words, directly optimizing the ROUGE metric helps improve performance almost as much as integrating syntactic information.
The combination of reinforcement learning with syntax information keeps increasing the score.
However, the resulting improvement is smaller than when adding syntax without RL.
We speculate that because the search space with syntax has a larger number of dimensions than without syntax, it may also be more difficult to explore with RL.

\paragraph{Parameters.} The baseline model has 6.617M parameters. This is increased by roughly 300K paramters for the Postag and the Deptag model and correspondingly by roughly 600K paramters for the Pos+Deptag model. In comparison, RL optimization does not involve any additional
parameters. However, it requires two more decoder passes for the sampling
and greedy predictions.

\paragraph{Speed.} Syntax-aware models are slightly longer to train than the baseline. Running on a single GPU GeForce GTX 1080, the baseline model requires 13h18m per epoch with 114k updates
while the training time of syntax-aware models increases by about 6\% (Postag s2s). Also, it takes one week to get the pre-processing tag labels of these syntactic features for the whole 3.8M training samples of Gigaword corpus on 16 cores cpu machine Dell Precision Tower 7810.
Surprisingly, adding the RL loss (which requires re-evaluating the ROUGE score for every generated text at every timestep) reduces training time by 12\%.
We speculate that the RL loss may act as a regularizer by smoothing the search space and help gradient descent to converge faster.

\begin{table*}[h]
\centering
\begin{tabular}{l|c|c|c|c|c|c|c|c|c|c}
\hline \bf Models & \multicolumn{4}{c|}{\bf Content words} & \multicolumn{5}{c|}{\bf Function words} & \bf MSE to gold \\ 
 & \bf NN & \bf VV & \bf JJ & \bf RB & \bf CD & \bf DT & \bf TO & \bf IN & \bf SYM & \bf to gold \\ \hline
Gold target & 49 & 12.5 & 12.9 & 1.6 &  1.3 & 1.5 & 2.5 & 10.6 & 4 & 0 \\
\hline
Our baseline s2s & 43.4 & 13.8 & 10.8 & 1.4 & 1.3 & 1.6 & 3.5 & 8.9 & 11.3 & 10.52 \\
\hline
Postag s2s & 50.4 & 14.5 & 12.1 & 1.3 & 1.3 & 1.1 & 3.9 & 8.1 & 2.1 & 2.07 \\
Deptag s2s & 49.8 & 14.5 & 12 & 1.7 & 1.3 & 1.1 & 3.7 & 8.9 & 1.9 & 1.59 \\
Pos+Deptag s2s & 51.4 & 14.7 & 12.6 & 1.6 & 1.1 & 1 & 3.7 & 7.9 & 1.8 & 2.72 \\
\hline
RL s2s & 50 & 14.1 & 11.9 & 1.3 & 1.6 & 1.2 & 4.1 & 9.1 & 1.6 & 1.71 \\
RL pos+deptag s2s & 49.9 & 14.3 & 12.4 & 1.3 & 1.5 & 1.2 & 3.6 & 9 & 2.2 & 1.28 \\
\hline
\end{tabular}
\caption{Proportion of generated postags}
\label{tab:prop_by_postag}
\end{table*}


\paragraph{Learning Curve.}

\begin{figure}
\centering
\includegraphics[scale=0.38]{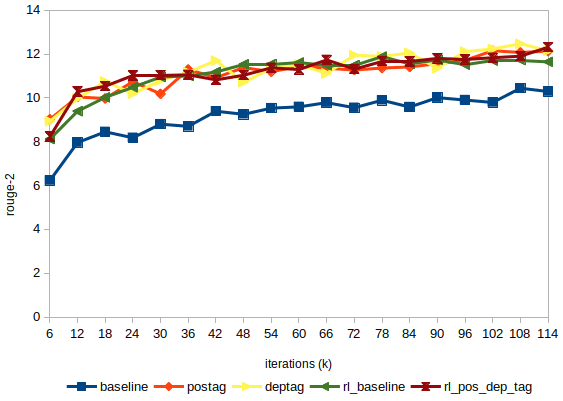}
\footnotesize
\caption{Evolution on test set during training}
\label{fig:learning_curve}
\end{figure}

Figure~\ref{fig:learning_curve} shows the evolution of Rouge-2 on the test set over 1 epoch.
We can observe that syntactic models obtain a better performance than the baseline after the first 6k iterations.
Sequence models with RL also quickly reach the same performance than syntactic models, even though the RL loss is only weakly taken into account at the start of training.
As learning continues, the gap between the top models (with syntax and/or RL) and the baseline stays constant.
The increased speed of training with RL, especially at the beginning of training, constitutes an important advantage in many experimental conditions.

\begin{figure}
\centering
\includegraphics[scale=0.50]{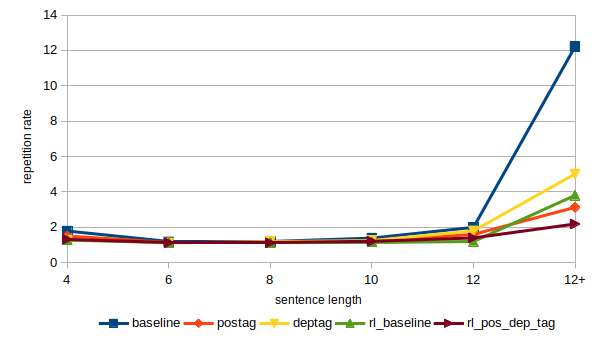}
\footnotesize
\caption{Repetition comparisons by length (lower is better)}
\label{fig:rep_by_length}
\end{figure}

\paragraph{Repetitions.}

Repetition is a common problem for sequence-to-sequence models~\cite{tu-etal-2016-modeling,Sankaran2016TemporalAM}. To assess the degree of repetitions in the generated summaries, we use~\cite{Le2017ImprovingST}'s repetition rate metric 
which is defined as follows:

\begin{equation}
\label{eq:rep_rate}
rep\_rat = \sum_{i=1}^{T(y)}\frac{1+r(\widetilde{y}_i)}{1+r(Y)}
\end{equation}
where $\widetilde{y}_i$ and $Y_i$ are the $i^{th}$ generated sentence
and $i^{th}$ gold abstract target sentence respectively, and $r$ is
the number of repeated words: $r(X) = len(X) - len(set(X))$.  $len(X)$
is the length of sentence $X$ and $len(set(X))$ is the number of words
that are not repeated in sentence $X$.
Figurer~\ref{fig:rep_by_length} compares the repetition rate of
several models; the horizontal axis is the length of sentences, and
the vertical axis is the repetition rate.
The proposed RL-model combined with syntactic information performs
the best on long sentences, with less repeated words than all
other models.
Indeed, short sentences are less likely to contain repetitions, but it is
interesting to observe that RL-training enriched with syntax improves
the quality of long sentences on this aspect.

\begin{figure}[t!]
\centering
\includegraphics[scale=0.53]{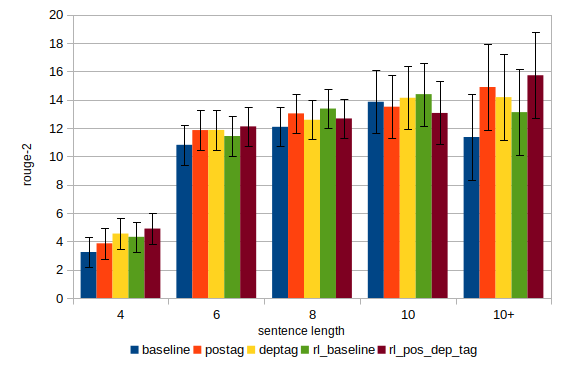}
\footnotesize
\caption{Performance comparisons by length}
\label{fig:rouge_by_length}
\end{figure}

\begin{table*}[h]
\small
\centering
\begin{tabular}{l|l}
\hline 
\bf Source & the us space shuttle atlantis separated from the orbiting russian mir space station early saturday , \\
& after three days of test runs for life in a future space facility , nasa announced .\\
\bf Abstract & atlantis mir part ways after three-day space collaboration by emmanuel unk \\
\bf Baseline s2s & \textit{atlantis atlantis atlantis separated from mir space station} \\
\hline
\bf Postag s2s & us space shuttle atlantis separated from mir space station \\
\bf Deptag s2s & atlantis separated from russian space station \\
\hline
\bf Rl s2s & us shuttle atlantis separated from mir \\
\bf Rl pos+deptag s2s & us shuttle atlantis separated from russian space station \\
\hline \hline
\bf Source & swedish telecommunications giant ericsson has reached a basic agreement to sell its relay production \\
& to japanese electronics company UNK corp , ericsson said thursday .\\
\bf Abstract & ericsson sells relay production to unk 's unk corp \\
\bf Baseline s2s & \textit{ericsson to sell its its to sell its} \\
\hline
\bf Postag s2s & ericsson to sell relay production to unk \\
\bf Deptag s2s & ericsson reaches basic agreement to sell relay production \\
\hline
\bf Rl s2s & ericsson reaches basic agreement to sell relay production \\
\bf Rl pos+deptag s2s & ericsson sells relay production to unk corp \\
\hline \hline
\bf Source & the shooting down of the largest transport plane in the sri lankan air force has wrecked supply lines \\
& and slowed a major government offensive against the tamil rebel citadel of jaffna , analysts said .\\
\bf Abstract & downing of plane slows sri lanka 's army onslaught on jaffna by amal jayasinghe \\
\bf Baseline s2s & \textit{sri lankan air force has} \\
\hline
\bf Postag s2s & sri lankan air force plane shooting down \\
\bf Deptag s2s & sri lanka 's largest transport plane shooting kills supply lines
 \\
\hline
\bf Rl s2s & sri lankan air force wrecked supply lines \\
\bf Rl pos+deptag s2s & sri lankan air force shooting down \\
\hline
\end{tabular}
\caption{\label{tab:examples_result} Generated examples of different models.}
\end{table*}

\begin{table}[t!]
\centering
\begin{tabular}{l|c|c}
\hline \bf Models & \bf Rel. & \bf Grammar. \\ \hline
Baseline s2s & 2.13 (+/-0.14) & 2.47 (+/-0.18) \\
\hline
Postag s2s & 3.26 (+/-0.18) & 4.19 (+/-0.16) \\
Deptag s2s & 3.17 (+/-0.19) & 4.2 (+/-0.17) \\
\hline
RL s2s & 3.23 (+/-0.19) & 4.26 (+/-0.16) \\
RL pos+deptag & 3.45 (+/-0.18) & 4.49 (+/-0.13) \\
\hline
\end{tabular}
\caption{Human Evaluations}\label{tab:human_eval}
\end{table}

\paragraph{Analysis by Postags.}
\label{ssect:anaysis_by_postags}

To further investigate the linguistic structure of the generated output, we compute for each POS tag class T, the proportion of POS tags of type T relative to the number of generated words (on the test set). We group POS tags into 9 classes: cardinal numbers (CD), determiners (DT), nouns and proper nouns (NN), verbs (VV), adjectives (JJ), adverbs (RB), to (TO), prepositions and subordinating conjunctions (IN) and symbols (SYM).

We evaluate whether the generated summary has a similar or different POS tags distribution than the ground truth by computing for each model
the mean square error (MSE) between every generated and gold POS tag class.
These errors are shown in Table~\ref{tab:prop_by_postag}.

On average and for all POS tag classes, both syntax-aware and RL models are much closer (about 5 times) to the gold than the baseline.
In a similar way as with repetitions, the best summarization model in terms of POS tag classes is the combined RL and syntax model.

\paragraph{Effects on Long Sentences.}

We group sentences of similar lengths together and compute the Rouge score.
Figure~\ref{fig:rouge_by_length} reports the Rouge-2 scores for various lengths of generated texts, with a 95\% \textit{t-distribution} confidence interval.
It shows that the RL and syntax models perform globally better than the baseline as sentences get longer.
For long sentences (more than 10 words), this effect is more pronounced, the syntax(+RL) models outperform significantly the RL and baseline models.

\paragraph{Human Evaluation.}

In order to evaluate the quality of the summaries produced by the models, we asked 3 annotators to assess the relevance and grammaticality quality of the summaries.
Each criterion is rated with a score from 0 (bad) to 5 (excellent).
Annotators are instructed to evaluate 50 samples randomly selected from the test set. The model information is anonymous to the annotators.
The evaluation results with a 95\% \textit{t-distribution} confidence interval is shown in Table~\ref{tab:human_eval}.
We see that RL performs on par with postag, deptag on relevance and grammaticality criterions and they all outperfom baseline. This is consistent with the results on POS tag classes above which indicate that these models generate more content words and less function words than the baseline.
Once again, RL with pos+deptag obtains the best result.

\paragraph{Qualitative Analysis.}

Table~\ref{tab:examples_result} shows some sample outputs of several models. 
Example 1 presents a typical repetition problem (the word \textit{``atlantis''}) often found in the baseline. Both syntax and RL models manage to avoid repetitions.
Example 2 shows that RL (without any syntactic information) can search and find surprisingly the same structure as the syntax-aware model.
In the last example, the baseline fails as it stops accidentally after a modal verb while syntax and RL models can successfully generate well-formed sentences with subject-verb-object.
However, semantically, RL and RL with pos+dep tag (like the baseline model) fail to capture the true meaning of the gold summary (\textit{``transport plane''}
instead of \textit{``air force''} should be the real subject in this case). Deptag s2s seems the best in terms of summarizing syntactic and semantic content on these examples.

\section{Conclusion}
We have studied in details in this work the quality of syntactic implication of the summaries that are generated by both syntactically-enriched summarization models and
reinforcement-learning trained models, beyond the traditional ROUGE-based metric classically used to evaluate summarization tasks.
We have thus focused on the quality of the generated summaries, in terms of the number of repeated words, which is a common issue with summarization systems,
but also in terms of the distribution of various types of words (through their POS-tags) as compared to the gold.
Because these aspects strongly depend on sentence length, we have also studied the impact of sentence length.
Finally, we have manually evaluated the quality of the generated sentences in terms of relevance and grammaticality.
Our results suggest that enriching summarization models with both syntactic information and RL training improves the quality of generation in all of these aspects.
Furthermore, when computational complexity is a concern, we have shown that RL-only models may be a good choice because they provide nearly as good results as
syntactic-aware models but with less parameters and faster convergence time.
We plan to extend this work by further applying similar qualitative evaluations to other types of summarization models and text generation tasks.

\section{ Acknowledgments}
This work has been funded by Lorraine Université d’Excellence; experiments realized in this work have been
done partly on Grid5000 Inria/Loria Nancy. We would like to thank all consortiums for giving us access to their resources.

\bibliography{aaai}
\bibliographystyle{aaai}

\end{document}